\documentclass[10pt,twocolumn,letterpaper]{article}

\usepackage{ijcb}
\usepackage{times}
\usepackage{epsfig}
\usepackage{graphicx}
\usepackage{amsmath}
\usepackage{amssymb}
\usepackage{multirow}
\usepackage{bm}

\usepackage{adjustbox}
\usepackage{setspace}
\usepackage[normalem]{ulem}

\usepackage[perpage]{footmisc}
\usepackage{booktabs}
\usepackage{subcaption}
\usepackage{setspace}
\usepackage{parselines}
\usepackage{lineno}
\usepackage{setspace}
\usepackage{hyperref}
\usepackage{emptypage}
\usepackage{rotating}
\usepackage{booktabs}
\usepackage{makecell}
\usepackage{algorithm,tabularx}
\usepackage{csquotes}
\usepackage{xhfill}

\ijcbfinalcopy 

\newcommand\bs\textbackslash 

\ifijcbfinal\fi

\begin{document}

\title{Robust Ensemble Morph Detection with Domain Generalization}


\author{Hossein Kashiani, Shoaib Meraj Sami, Sobhan Soleymani, Nasser M. Nasrabadi\\
West Virginia University\\
{\tt\small \{hk00014,sms00052, ssoleyma\}@mix.wvu.edu, nasser.nasrabadi@mail.wvu.edu
}}

\maketitle
\thispagestyle{empty}

\begin{abstract}

 Although a substantial amount of studies is dedicated to morph detection, most of them fail to generalize for morph faces outside of their training  paradigm. Moreover, recent morph detection methods are highly vulnerable to adversarial attacks. In this paper, we intend to learn a morph detection model with high generalization to a wide range of morphing attacks and high robustness against different adversarial attacks. To this aim, we develop an ensemble of convolutional neural networks (CNNs) and Transformer models to benefit from their capabilities simultaneously. To improve the robust accuracy of the ensemble model, we employ  multi-perturbation adversarial training and generate adversarial examples with high transferability for several single models. Our exhaustive evaluations demonstrate that the proposed robust ensemble model generalizes to several morphing attacks and face datasets. In addition, we validate that our robust ensemble model gain better robustness against several adversarial attacks while outperforming the state-of-the-art studies.

\end{abstract}

\begin{figure}[t] 
\centering
    \includegraphics[width=0.47\textwidth]{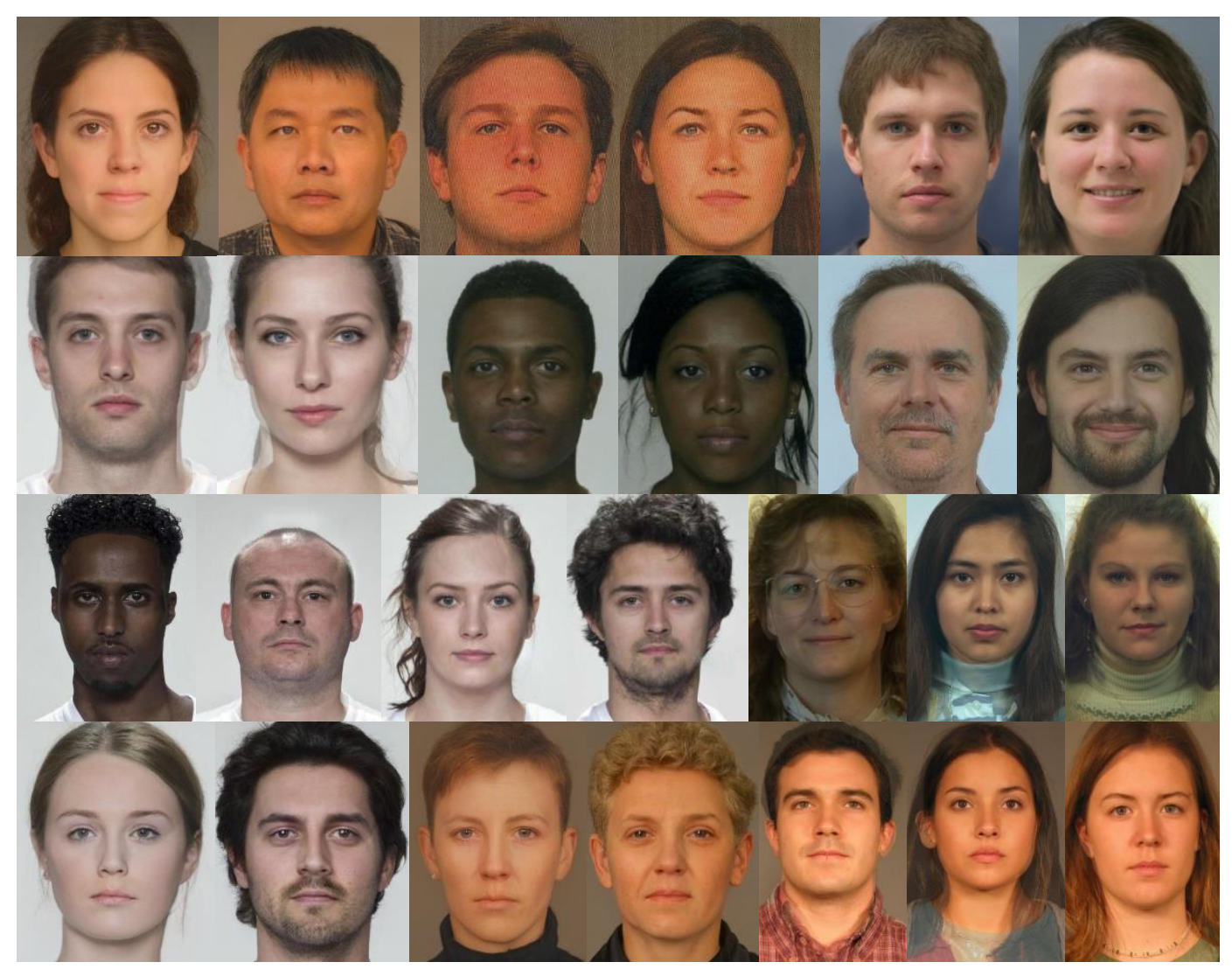}
    \caption{Domain shift between different morphing attacks in different datasets, including  FERET \cite{FERET}, FRLL \cite{debruine2017face,Sarkar2020}, FRGC \cite{phillips2005overview}, {and} AMSL \cite{neubert2018extended}.  {Morphs} correspond to  landmark-based morphing {methods} such as Facemorpher \cite{facemorpher_2019} and WebMorph \cite{WebMorph} {and} GAN-based morphing {methods} such as MIPGAN \cite{zhang2021mipgan} and StyleGAN2 \cite{Sarkar2020}.}
    \label{fig1}
\end{figure}

\section{Introduction}

Face recognition systems are {built upon} the hypothesis that {the} face is  {uniquely linked} to  {the} identity. When it comes to face verification scenarios,  morph attack{s are} potential threat{s} due to {their} capacity to break this unique connection. Face morph attacks are particularly dangerous for border security since a morphed passport photo allows unauthorized entry {to be} unnoticed. Through this loophole, a criminal can {apply} for a passport using a morphed face. Face morphing is an image manipulation technique in which two {faces} are blended together to form a morphed image of high quality. At the time of passport {enrollment}, the passport photo can be easily manipulated with a morphing attack without the requirement of advanced passport {forgery technologies}. {Therefore}, face recognition systems rely highly on the morph attack detection. 

A typical assumption in morph detection studies \cite{scherhag2017vulnerability,chaudhary2021differential,aghdaie2021detection,aghdaie2021attention} is that the source and target data are independent and identically distributed ({\it i.i.d.}) \cite{soleymani2021mutual,banerjee2021conditional,damer2019detect}. This {assumption} neglects the domain shift challenge in the real world (Figure \ref{fig1}). As a result, when applied to out-of-distribution {samples}, a learning agent that {is} only trained {on} source data often performs much worse. The domain shift issue may be mitigated to some extent by {including} some data from the target domain, but this is a non-trivial task {since} preparing image annotations is both costly and time‐consuming. Moreover, despite the remarkable success of deep CNNs in face recognition systems, they are known to be susceptible to another type of attack known as adversarial attack. An adversarial attack attempts to generate an adversarial example with indistinguishable and purposeful perturbations, forcing CNNs to an incorrect prediction. In this regard, the vulnerability of morph detection systems to adversarial examples have serious consequences.

In this study, we develop an ensemble  {of} deep models trained on a morph source domain to improve domain generalization. It should be noted that neural networks mostly do not attain perfect performance owing to the overwhelming number of local minima. In other words, neural networks might fall into a variety of local minima. As a result, individual neural networks do not operate optimally in various areas of the feature space, and the errors do not have a large positive correlation between them \cite{ganaie2021ensemble}. To address this issue, we adopt {an} ensemble of well-trained neural network{s}. Our evaluations demonstrate that the ensemble learning can partly mitigate the domain shift challenge. In this respect, we {study} an ensemble model, which comprises of a ViT B-16 Transformer~\cite{dosovitskiy2020image}, ViT L-32 Transformer~\cite{dosovitskiy2020image}, and a noise-aware Inception ResNet network~\cite{he2016deep,venkatesh2020detecting}. Noise-aware Inception ResNet network is {built upon Venkatesh {\it{et al.}},~}\cite{venkatesh2020detecting} with {an} Inception ResNet backbone. Domain generalization needs translation equivariance, local receptive fields, weight sharing, and long-range dependencies. Integration of CNN and Transformer models can incorporate these objectives at the same time. Therefore, we propose an ensemble model to include the strengths of both CNN and Transformer architectures  \cite{venkatesh2020detecting,he2016deep,dosovitskiy2020image,tan2019efficientnet, huang2019ccnet} simultaneously.

In addition, to robustify our ensemble model against adversarial perturbations, we first generate adversarial examples with high transferability in a black-box setting. To this end, we generate a wide range of adversarial perturbations with different methods such that our adversarial examples fool different morph detection methods in a white-box setting. Then, {we robustify our ensemble model using multi-perturbation adversarial training \cite{goodfellow2014explaining} }. With the multi-perturbation adversarial training, we enforce the output of the proposed ensemble model {to} remain nearly within an $l_{p}$ ball of every training {sample} to enhance robust accuracy while keeping favorable clean accuracy. Our major contributions in this papers are:
\begin{itemize}
    \item We integrate CNN and Transformer models and propose an ensemble model for morph detection that highly generalizes to a wide range of morphing attacks.
    \item We craft highly transferable adversarial examples for multi-perturbation adversarial training to improve the adversarial robustness of our ensemble models.
    \item We carry out extensive evaluations on different datasets to prove the generalization capability and adversarial robustness of our ensemble model.

\end{itemize}

\section{Related Works}
\subsection{Morph Detection}
Morph attack detection methods are divided into single and differential morph detection. In the single morph detection, the detector only aims to detect the potential morphed image for final classification. This corresponds to morph detection during the passport application, when the applicant uploads his passport picture. On the other hand, differential morph detection utilizes an additional trusted image of the real subject for its detection. To be more specific, it makes a comparison between the potential morphed image and the trusted reference image.

Recently, deep learning models have been widely used for morph detection. Authors in \cite{banerjee2021conditional} employ a conditional generative network (cGAN) for differential face morph attack detection. The cGAN learns to implicitly extract identities from the morphed image conditioned on the trusted reference image. Aghdaie {\it et al.} \cite{aghdaie2022morph} exploit wavelet domain analysis to explore spatial frequency information for morph detection. They train a CNN-based morph classifier on the decomposed wavelet sub-bands of the morphed and bona fide images. Authors in \cite{venkatesh2020detecting} propose a deep multi-scale context aggregation model to capture residual morphing noise. The pre-trained AlexNet \cite{krizhevsky2012imagenet} is then utilized to compute the textural features of the residual morphing noise for morph attack detection. Recently, Damer {\it et al.} \cite{damer2022privacy} introduce a synthetic-based development dataset for morph attack detection to address the legal issues of utilizing biometric information.

\begin{figure*}[!t]
\centering
    \includegraphics[width=0.99\textwidth]{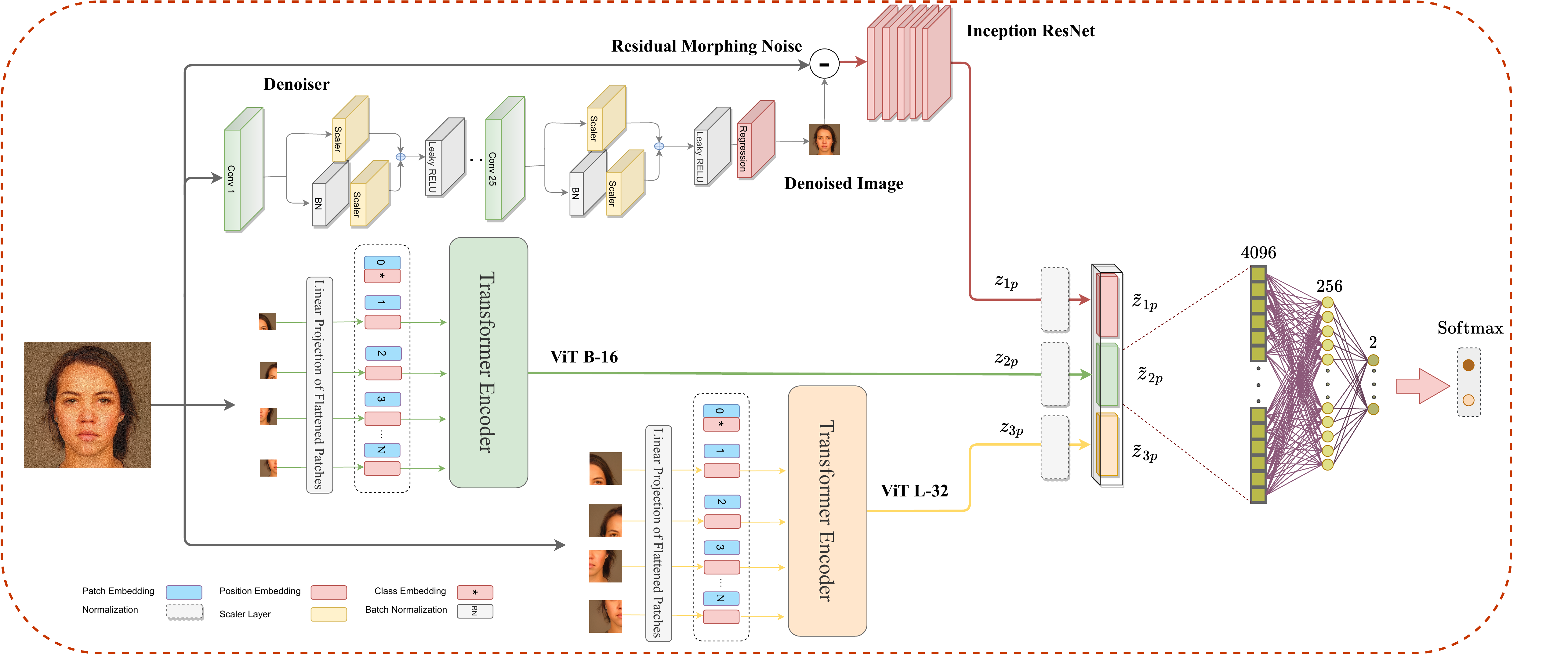}
    \caption{ Our proposed ensemble model consists of three single models: ViT B-16 Transformer, ViT L-32 Transformer, and noise-aware ResNet. The single models is pre-trained on morphed face images.}
    \label{fig2}
\end{figure*}
\subsection{Domain Generalization and Ensemble models}
Machine learning paradigms \cite{nodehi2021multi,mostofa2021deep,alipour2021superpixel,nourelahi2022machine,mosharafian2021gaussian} are built upon the assumption that training and test data have the same probability distributions. When this hypothesis is even marginally broken, as most real-life settings, a significant drop in performance can be observed. Domain generalization aims to address this issue and mainly covers the events in which the target source is unavailable during the training phase. In this respect, ensemble learning is investigated for domain generalization. Domain-specific neural networks \cite{zhou2021domain}, and weight averaging \cite{cha2021domain} are a number of domain generalization-based ensemble learning studies. In ensemble learning, we seek to learn several instances of the same model with different initializations, or we learn different explicit models. In this respect, Xu {\it et al.} \cite{xu2021how} demonstrate that naively integrating diverse training distributions from different source domains can contribute to out-of-distribution generalization. 
 In our study, domain generalization from a single source domain is targeted using different state-of-the-art explicit deep learning models. We construct our ensemble model from CNN and Transformer models since the integration of these models enables us to capture long-term and short-term connections in the visual data.

\subsection{Adversarial Robustness}
While face recognition is improved significantly over recent years thanks to the breakthroughs in deep learning, the face recognition models are prone to adversarial perturbations \cite{zhong2020towards}. This shortcoming is severe since certain adversarial perturbations created by the white-box attacks can fool other unrelated models owing to the adversarial transferability phenomenon \cite{huang2019enhancing}. One of the most thoroughly benchmarked defenses against adversarial attacks is adversarial training \cite{tramer2019adversarial,cui2021learnable}, in which a CNN model is trained on adversarial examples generated by itself. Despite the considerable improvement in adversarial learning, recent works on biometrics mostly rest on developing robust face recognition \cite{zhong2020towards}, face identification \cite{dong2019efficient}, and face verification systems \cite{kilany2021analysis}. However, in the case of the morphed images, it has received far less attention to generate adversarial examples and benchmark their robustness against adversarial perturbations. To alleviate this threat, we augment the training dataset of the target model with multi-adversarial perturbations transferred from different pre-trained models. 

\section{ {Proposed} Framework}
Ensemble learning integrates {several} single models to boost {the} performance, improve generalization, and mitigate the overfitting challenge in comparison to base model{s}. A single model can {converge to} a {local} minima. However, an ensemble {of single} model{s} with different initialized weights, individual loss functions, and individual gradient backpropagation{s} {can} provide a guaranteed and faster convergence to the global minima \cite{ganaie2021ensemble}. This is because {different} single models {usually} do not make the same mistakes. Inspired by ensemble learning, we construct our model from the state-of-the-art individual CNN and Transformer models. The CNNs are built with the inherent inductive bias of locality. The scale-invariance can also be provided in CNNs by the multi-scale features, which can be computed by varying kernel sizes and dilation rates in the intra- or inter-layer convolutions of the CNN hierarchical structure \cite{xu2021vitae}. Furthermore, CNNs take into account local correlations among neighboring pixels in order to capture local features, which results in the locality attribute. While Transformer models lack precious inductive biases in visual feature extraction, they model long-range dependencies in visual data. To incorporate the inductive bias and model long-range dependencies, we build our ensemble model by means of integrating the Transformer and CNN models in {an ensemble model }as depicted in Figure \ref{fig2}.

\subsection{Inception-ResNet Morph Classifier}
The {RGB} residual morphing noise {is} demonstrated to be effective for morph detection \cite{venkatesh2020detecting}. 
Residual morphing noise usually exists in high-frequency patterns of the morphed face images. The high-frequency patterns are the fine irregularities in the morphed face images that can model the morphed manipulation and be useful for morph detection. As the morphing artifacts are random and non-deterministic, we learn a CNN denoiser to calculate the residual artifacts for morph attack detection task. The CNN denoiser is composed of 15 dilated convolutions with $3 \times 3$ kernel sizes and growing dilation factors as \cite{venkatesh2020detecting} for multi-scale contextual information aggregation. At each stage, the output of each convolution is divided into two branches: an identity branch with a scale layer and a batch normalization branch. These two branches are then combined together and fed to a leaky RELU and an adaptive normalization \cite{chen2017fast} to finally calculate the residual morphing artifacts. Given the calculated residual morphing artifacts, we utilize the Inception-ResNet-V1 model \cite{szegedy2017inception} {for} morph detection. In this paper, we call this model the noise-aware ResNet model.

\subsection{Transformer Classifier}
Due to their capability in capturing long-term dependencies between different regions of the data, Transformers are considered in many tasks such as object recognition, image classification, and face recognition. Morph detection problem can also benefit from this capability since long-term dependencies between different regions in facial images can provide important information. We leverage from the ViT B-16 and ViT L-32 \cite{dosovitskiy2020image} Transformers for morph detection task in our ensemble model. 
\paragraph{Tokenization}
A ViT Transformer applies attention to small patches of input images instead of pixels. To be more specific, ViT divides a 2D image $ x \in \mathbb{R} ^{H \times W \times 3}  $ into $ R \times R $ flattened patches $ x _{f} \in \mathbb{R} ^{N \times(R^2 \times 3)}$ uniformly, where {(}$ H \times W ${)}, $ R ^{2} $, 3, and  $ N $  stand for spatial dimensions of the input image and patches, three RGB channels, and the number of total patches{, respectively}. It is worth noting that ViT {underperforms for} both extremely small and large patch sizes since the former induce too much computational cost, while the latter cannot encode the low-level context in the input images. The first layer of ViT projects the flattened patches (also called patch tokens) into latent D-dimensional embeddings \cite{dosovitskiy2020image}. To learn about the structure of the input data, a positional embedding is added to the flattened patches {which} contains information about the location of patches. An extra classification token  $ x _{class} $ is also appended to the embedded patches. The content of {this classification} token interacts with all the remaining patch tokens in the self-attention module, accumulating data for the eventual classification \cite{dosovitskiy2020image}. With this structure, nearby patches contain more comparable position embeddings, and the embeddings of patches in the same row or column are identical. Thus, the ViT utilizes the latent position embedding to consider distances within the input image patches. The resulting vector corresponding to different patch tokens is then fed into the transformer encoders. Note that the $ x _{class} $  in the Transformer encoder finally results in $ y $ output \cite{dosovitskiy2020image}.

\paragraph{Encoder Block{:}} Each block of a Transformer encoder is composed of a feed-forward network (FFN) and a multi-headed self-attention module (MSA). The hidden layer of the FFN is constructed from a two-layer multi-layer perceptron as well as a {Gaussian Error Linear Unit (GELU)} non-linearity layer. Around each sub-layer, residual shortcuts are adopted, which are preceded by a layer normalization (LN). The encoder of ViT works as follows:
\begin{equation}\label{eq.1}
y = \hat{x} + \mathrm{FFN}(\mathrm{LN}(\hat{x})),
\end{equation}
\begin{equation}\label{eq.2}
 \hat{x} = x + \mathrm{MSA}(\mathrm{LN}(x)),
\end{equation}
\noindent where $x$ and $y$ are the out of tokenization phase and the final output class, respectively. A series of encoders component constitutes a Transformer encoder.
\paragraph{Multi-Head Self-Attention{:}} Multi-head attention extends single-head self-attention to encompass various complicated connections among distinct image patches. The self-attention mechanism is a fundamental building block of Transformers that systematically simulates the relationship between all pixels in an image by giving a pairwise attention score between every two patch tokens in {the} terms of global contextual information. To this end, the embeddings of $R\times R$ pixel patches, $ x _{f} $, are mapped to three learnable matrices: query ($ \textbf{Q}  \in \mathbb{R} ^{ N+1 \times C  }$), key  ($ \textbf{K}  \in \mathbb{R} ^{ N+1 \times C } $) , and value ($ \textbf{V}  \in \mathbb{R} ^{ N+1 \times C  }$). Then{,} the self-attention mechanism {is defined as}:

\begin{equation}\label{eq.3}
 \mathrm{Attention} (\textbf{Q,K,V}) = \mathrm{softmax}( \frac{\textbf{Q}\textbf{K}^\textbf{T}}{\sqrt{d_k}}) \,\textbf{V},
\end{equation}
\noindent where $d_k$ is the dimension of $K$ matrix. This single self-attention process is performed for each head  in parallel in the MSA module. Then, the output of various head units is concatenated along the channel dimension.

\subsection{Fusion}
  \noindent {To construct our ensemble model, we train a FFN to compute the matching scores of all single models in the fusion phase. For this goal, we capture score $z_i$ from the output of each model $i \in {1,2,3}$. Since the large scores which belong to a certain model can dominate the others and force them to be small, the scores of single models are first normalized as follows :} 
  
\begin{equation}\label{eq.4}
\widetilde{z}_{ip }=\frac{e^{z_{ip}}}{\sum_{j} e^{z_{ij}}}, \: \: i = 1,2,3,
\end{equation}
  
 \noindent where $\widetilde{z}_{ip}$ is the normalized score of a single model. Finally, the normalized scores of all single models are ensembled for final classification as follows:
 
\begin{equation}\label{eq.4}
y= \mathrm{FFN}[ \widetilde{z}_{1p}; ...;  \widetilde{z}_{ip}], 
\end{equation}

\noindent where FFN denotes a two-layer feed forward network with hidden layers of $256$ and $2$ units. The FC layers are also activated with {a} RELU function. Note that{, during the training,} we randomly set a limited number of {these} scores to zero to cope with the overfitting.

\subsection{Multi-Perturbation Adversarial Training}
Adversarial training is one of the main types of defense against adversarial attacks that aims to train moderately robust DNNs with adversarial examples \cite{tramer2019adversarial}. In adversarial training, we require to delicately set the perturbation level $\epsilon$ for the generation of adversarial examples in the training phase. Using small values of $\epsilon$ in adversarial training fails to generate hard adversarial examples, and consequently, the final trained model would lack adversarial robustness. On the other side, large values of $\epsilon$ would improve the adversarial robustness at the cost of a significant clean accuracy drop. As such, we need to tune the perturbation level $\epsilon$ to strike the right balance between robust and clean accuracy. Moreover, the adversarially trained models with a certain type of adversarial perturbation are still vulnerable to other types of adversarial attacks \cite{tramer2019adversarial}. To cope with this issue, we adopt ensemble-based multi-perturbation adversarial training. To this end, we augment training data with a wide range of adversarial perturbations. In a white-box setting, we fool each single models and generate adversarial examples with different perturbation level $\epsilon $. It is worth highlighting that several types of adversarial attacks are included in order to improve the network robustness. 

To improve the effectiveness of our adversarial training, we require to craft the adversarial examples with high transferability. Conventionally, the adversarial examples can fool the white-box models more easily compared to the black-box models. The adversarial perturbations generally do not transfer well to the black-box models with different topologies \cite{huang2019enhancing}. Transferability, which allows adversarial examples to transfer to networks of unknown structures, makes adversarial examples even more harmful. To address this challenge and generate highly transferable adversarial examples, {we utilize} the model-based ensembling Attack \cite{liu2017delving}. In this approach, we assume that if an adversarial example succeeds in fooling several models, it can also fool black-box models with high probability \cite{liu2017delving}. With that in mind, we produce adversarial examples for an ensemble of single models in a white-box setting by {optimizing the following equation}:

\begin{equation}
\operatorname{argmax}_{x ^\star}-\log \left(\left(\sum_{i=1}^{n} \alpha_{i} J_{i}\left(x^{\star}\right)\right) \cdot 1_{y}\right)+\lambda d\left(x, x^{\star}\right),\label{eq_op}
\end{equation}

\noindent where $x$, $y$, and $n$ denote the input image, the {ground-truth} class, and the number of single models. {In addition}, $J_i$, $\alpha_{i}$ and $d$ indicate the output of a single model, ensemble weights, and the distance between input and perturbed {images}. This optimization results in the alignment of the decision boundaries of the individual models. Therefore, the crafted adversarial example would have higher transferability in the black-box setting. Equipped with these highly transferable adversarial examples, we employ multi-perturbation adversarial training over single models. To be more specific, we apply this optimization (Equation \ref{eq_op}) to several untargeted adversarial attacks to craft highly transferable adversarial images and utilize them in the adversarial training of the ensemble model. These attacks include fast gradient sign method (FGSM) \cite{43405}, basic interactive method (BIM) \cite{45818}, Random initialization FGSM (RFGSM) \cite{2018ensemble}, projected gradient method (PGD), \cite{madry2018towards}, PGDL \cite{madry2018towards}, Trade-off PGD (TPGD) \cite{zhang2019theoretically}, Auto-PGD (APGD) \cite{croce2020reliable}, APGD-Targeted (APGDT) \cite{croce2020reliable}, SmoothFool \cite{dabouei2020smoothfool}, and AutoAttack \cite{croce2020reliable}.

\section{Evaluations}

We benchmark the generalization performance of the proposed ensemble model on a wide range of unseen morph attacks. In a real-world scenario, the morphing technology used by a criminal attacker may be novel and unknown. As such, it is reasonable to evaluate morph detection methods on several "unseen" datasets to obtain an accurate representation of how the network performs. To this end, a wide range of target domains are employed in our investigations. To assess network performance, we rely on the Area Under the Receiver Operating Characteristic Curve (AUC), Attack Presentation Classification Error Rate (APCER), and  Bona fide Presentation Classification Error Rate (BPCER), which are the standard measures for morph attack detection. APCER stands for Attack Presentation Classification Error Rate or the rate at which morphs are incorrectly categorized as bona fides. BPCER, on the other hand, is the rate at which bona fide images are wrongly identified as morphs \cite{soleymani2021mutual}.

\subsection{Experimental setup}
To train the single models, the input images are first pre-processed with the MTCNN framework \cite{zhang2016joint}. Faces are detected, aligned, and resized to 512 × 512 using the MTCNN framework. All training images are further augmented with horizontal flips. Additionally, the batch generator is weighted to mitigate the class imbalance at every iteration. We utilize our private dataset as the source domain, which includes about 9,052 bona fide and 1,2991 morphed images. The target domains consist of five separate datasets, including FERET \cite{FERET}, FRLL \cite{debruine2017face,Sarkar2020}, FRGC \cite{phillips2005overview}, and AMSL \cite{neubert2018extended} datasets. To train the base single models on the source domain, we use the mini-batch stochastic gradient descent optimization with  batch size 256, and the Adam solver. The initial learning rate is set to $5 \times 10^{-5}$, and it is decreased by a factor of 2 at {$20^{th}$ and $30^{th}$} epochs. In {the} adversarial training, to strike a right robustness/accuracy balance, the adversarial examples are crafted with $\epsilon = {2}/{255}$ and $\epsilon = {4}/{255}$ perturbation levels. In the test phase, the adversarial attacks are produced in a white-box and black-box settings with translation-invariant FGSM (TIFGSM) \cite{dong2019evading}, Carlini and Wagner (C\&W) \cite{7958570}, momentum
iterative FGSM (MIFGSM) \cite{dong2018boosting}, Square \cite{andriushchenko2020square},  Diverse Inputs
Iterative FGSM (DIFGSM) \cite{xie2019improving}. In the black-box settings, the adversarial images are crafted using a black-box model and transfer to the ensemble model. Our experiments in the training and tests phases are conducted on three 12 GB TITAN X (Pascal) GPUs.



\subsection{Results and Analysis}

To explore the generalization capability of our ensemble model on a wide range of domains, we employ FERET \cite{FERET}, FRLL \cite{debruine2017face,Sarkar2020}, FRGC \cite{phillips2005overview}, and AMSL \cite{neubert2018extended} datasets in our evaluations. Further, different landmark-based and GAN-based morphing attacks are used to generate morph images on the datasets. The landmark-based attacks include Facemorpher \cite{Sarkar2020}, OpenCV \cite{Sarkar2020}, and WebMorph \cite{Sarkar2020} and GAN-based attacks include MIPGAN \cite{zhang2021mipgan}, StyleGAN2 \cite{Sarkar2020}, and Print and Scan attacks \cite{zhang2021mipgan}. Table \ref{dataset} presents the details of different test sets that are used in our experiments.

\begin{table}[!t]
\center
\caption{Test datasets utilized in our experiments.}
\resizebox{0.45\textwidth}{!}{%
\begin{tabular}{cccccc}
\toprule
\toprule

& Dataset Number&Name& Morph Images &Bona fide Images\\ \midrule
  &1&AMSL \cite{neubert2018extended}&2175&204\\
   \addlinespace[1mm]
 &2&FRLL AMSL \cite{Sarkar2020} &6525&204\\
  
 \addlinespace[1mm]
&3&FRLL Webmorpher \cite{Sarkar2020}&1221&  204\\
 \addlinespace[1mm]
&4&FRLL OpenCV \cite{Sarkar2020}&1221&204\\
 \addlinespace[1mm]
 &5&FRLL StyleGAN \cite{Sarkar2020} &1221&204\\
 \addlinespace[1mm]
 &6&FRLL Facemorpher \cite{Sarkar2020}&1221&204\\
 \addlinespace[1mm]
 &7&FERET OpenCV \cite{Sarkar2020}&529&1413\\
 \addlinespace[1mm]
  &8&FERET StyleGAN \cite{Sarkar2020}&529&1413\\
 \addlinespace[1mm]
  &9&FERET Facemorpher \cite{Sarkar2020}&529&1413\\
     \addlinespace[1mm]

  &10&FRGC OpenCV  \cite{Sarkar2020}&964&3038\\
     \addlinespace[1mm]
  &11&FRGC StyleGAN  \cite{Sarkar2020}&964&3038\\
   \addlinespace[1mm]
  &12&FRGC Facemorpher  \cite{Sarkar2020}&964&3038\\
   \addlinespace[1mm]
  &13&FRGC MIPGAN  \cite{zhang2021mipgan}&747&373\\
    \addlinespace[1mm]
  &14&FRGC MIPGAN + PRINT AND SCAN  \cite{zhang2021mipgan}&747&376\\

 \addlinespace[1mm]
\bottomrule
\bottomrule
\end{tabular}\label{dataset}
}

\end{table}%

\begin{table*}[!t]
\center
\caption{Morph detection {results} for different single models in terms of AUC, APCER (@BPCER=1\%), and BPCER (@APCER=1\%) metrics. N-ResNet denotes the noise-aware ResNet method \cite{venkatesh2020detecting}.}
\resizebox{.99\textwidth}{!}{%
\begin{tabular}{ccccccccccccccccccc}
\toprule
\toprule

   && Methods  & {\rotatebox[origin=c]{60}{\makecell{AMSL} }}& 
   {\rotatebox[origin=c]{60}{\makecell{FRLL \\AMSL} }}& 
   {\rotatebox[origin=c]{60}{\makecell{FRLL\\ Webmorpher  }}}&
   {\rotatebox[origin=c]{60}{\makecell{FRLL \\OpenCV }}}&
   {\rotatebox[origin=c]{60}{\makecell{FRLL \\StyleGAN}}}&
   {\rotatebox[origin=c]{60}{\makecell{FRLL\\ Facemorpher }}}&
   {\rotatebox[origin=c]{60}{\makecell{FERET\\ OpenCV}}}& 
   {\rotatebox[origin=c]{60}{\makecell{FERET\\ StyleGAN}}}&
   {\rotatebox[origin=c]{60}{\makecell{FERET\\ Facemorpher}}}&
   {\rotatebox[origin=c]{60}{\makecell{FRGC \\OpenCV}}}& 
   {\rotatebox[origin=c]{60}{\makecell{FRGC \\StyleGAN}}}&
   {\rotatebox[origin=c]{60}{\makecell{FRGC \\Facemorpher}}}&
   {\rotatebox[origin=c]{60}{\makecell{FRGC \\MIPGAN}}}&
   {\rotatebox[origin=c]{60}{\makecell{FRGC \\MIPGAN\\ (PRINT)}}}\\ \midrule
 

 \addlinespace[1mm]
 \addlinespace[1mm]

  \multirow{7}{*}{\rotatebox[origin=l]{90}{\makecell{AUC}}}

  &&ResNet  \cite{he2016deep}&99.32 & 98.77 &79.09& 99.99 &99.16 &99.94
&94.14&92.7&94.31&97.24&92.28&96.36&81.24&68.45&\\
 \addlinespace[1mm]

  &&N-ResNet  \cite{he2016deep,venkatesh2020detecting}& 99.63 &99.46 &87.4&99.95 &97.71 &99.83&94.51&95.02&95.88&99.65&97.07&99.61&91.68&73.91&\\
 \addlinespace[1mm]

&&EfficientNet  \cite{tan2019efficientnet, huang2019ccnet}& 99.95
 &99.73
& 83.52
 &99.85
 &90.71
&99.80
&89.01
&90.88
&92.97
&92.64
&68.75
&92.07
&95.66
&75.87
&\\
 \addlinespace[1mm]

  &&ViT B-16 \cite{dosovitskiy2020image}& 99.62 & 99.41 &90.70& 99.84 &86.37 &99.61&93.35&90.77&93.38&99.01&87.77&97.54&88.75&83.56&\\
 \addlinespace[1mm]

  &&ViT L-32\cite{dosovitskiy2020image}& 99.20 & 99.09 &85.81& 99.69 &93.38 &99.52&92.76&91.97&92.69&98.75&89.46&97.31&80.00&82.16&\\

 \addlinespace[1mm]
 \addlinespace[1mm]
\specialrule{.07em}{.1em}{.1em}
 \addlinespace[1mm]
  \addlinespace[1mm]
 \multirow{7}{*}{\rotatebox[origin=c]{90}{\makecell{APCER}}}

  &&ResNet  \cite{he2016deep}& 6.86
 & 8.82
 &72.54
& 00.0&7.35
 &00.0&43.94
&50.74
&43.38
&17.70
&39.4
&21.95
&84.182
&91.48
&\\
  
 \addlinespace[1mm]
  &&N-ResNet  \cite{he2016deep,venkatesh2020detecting}& 6.37 & 8.33 &76.96& 00.0&24.01 &1.96&54.91
&36.51&43.94&2.40&23.86&2.50&63.80&62.23&\\
  
 \addlinespace[1mm]
  &&EfficientNet  \cite{tan2019efficientnet, huang2019ccnet}& 0.98
 & 3.43
 &58.33
& 1.47
&53.92
 &1.47
&67.72
&55.41
&53.07
&37.12
&61.68
&39.10
&30.29
&82.71
&\\
  
 \addlinespace[1mm]
  &&ViT B-16 \cite{dosovitskiy2020image}&5.74 & 8.85 &75.51& 1.47&76.92 &4.09&39.50&66.91&38.75&15.04&74.68&29.77&70.95&74.43&\\
  
 \addlinespace[1mm]

  &&ViT L-32 \cite{dosovitskiy2020image}& 8.78
 & 8.70 &80.75& 2.04&49.09 &4.50&63.13&64.83&65.97&21.16&76.55&36.41&76.17&89.29&\\
  
\addlinespace[1mm]
 \addlinespace[1mm]
\specialrule{.07em}{.1em}{.1em}
 \addlinespace[1mm]
  \addlinespace[1mm]
 \multirow{7}{*}{\rotatebox[origin=c]{90}{\makecell{BPCER}}}

  &&ResNet  \cite{he2016deep}& 24.55
 & 36.78
 &98.44
& 0.32
&27.00
 &0.40
&44.23
&68.62
&46.88
&81.63
&93.98
&89.73
&98.39
&98.25
&\\
  
 \addlinespace[1mm]
&&N-ResNet  \cite{he2016deep,venkatesh2020detecting}&5.97
& 7.75
&70.35
& 0.0
&30.76
&1.71
&39.69
&61.62
&35.91
&14.00
&70.33
&16.39
&94.64
&99.59
&\\
  
 \addlinespace[1mm]
  &&EfficientNet  \cite{tan2019efficientnet, huang2019ccnet}
& 1.79
&8.93
& 95.08
&2.62
&72.42
&2.86
&65.97
&71.64
&57.08
&79.56
&99.37
&80.80
&93.17
&90.62
&\\
  
 \addlinespace[1mm]
  &&ViT B-16 \cite{dosovitskiy2020image}
&6.86
&6.86
&55.39
&2.94
&77.94
&6.37
&61.07
&66.31
&56.40
&17.281
&69.32
&31.764
&71.04
&89.89
&\\
  
 \addlinespace[1mm]

&&ViT L-32 \cite{dosovitskiy2020image}
&13.72
&20.09
&69.11
&6.37
&57.84
&9.80
&44.16
&62.42
&40.90
&20.17
&70.53
&41.70
&89.81
&86.43
&\\

 \addlinespace[1mm]

 \addlinespace[1mm]
\bottomrule
\bottomrule
\end{tabular}%
}

\label{result_1}%
\end{table*}%

\begin{table*}[!t]
\center
\caption{Morph detection results for different combinations of ensemble models in terms of APCER (@BPCER=1\%) and AUC metrics. N-ResNet denotes the noise-aware ResNet method \cite{venkatesh2020detecting}.}
\resizebox{.99\textwidth}{!}{%
\begin{tabular}{ccccccccccccccccccc}
\toprule
\toprule

   && Methods  & {\rotatebox[origin=c]{60}{\makecell{AMSL} }}& 
   {\rotatebox[origin=c]{60}{\makecell{FRLL \\AMSL} }}& 
   {\rotatebox[origin=c]{60}{\makecell{FRLL\\ Webmorpher  }}}&
   {\rotatebox[origin=c]{60}{\makecell{FRLL \\OpenCV }}}&
   {\rotatebox[origin=c]{60}{\makecell{FRLL \\StyleGAN}}}&
   {\rotatebox[origin=c]{60}{\makecell{FRLL\\ Facemorpher }}}&
   {\rotatebox[origin=c]{60}{\makecell{FERET\\ OpenCV}}}& 
   {\rotatebox[origin=c]{60}{\makecell{FERET\\ StyleGAN}}}&
   {\rotatebox[origin=c]{60}{\makecell{FERET\\ Facemorpher}}}&
   {\rotatebox[origin=c]{60}{\makecell{FRGC \\OpenCV}}}& 
   {\rotatebox[origin=c]{60}{\makecell{FRGC \\StyleGAN}}}&
   {\rotatebox[origin=c]{60}{\makecell{FRGC \\Facemorpher}}}&
   {\rotatebox[origin=c]{60}{\makecell{FRGC \\MIPGAN}}}&
   {\rotatebox[origin=c]{60}{\makecell{FRGC \\MIPGAN\\ (PRINT)}}}\\ \midrule
   
 \addlinespace[1mm]
 \addlinespace[1mm]

  \multirow{12}{*}{\rotatebox[origin=l]{90}{\makecell{AUC}}}

  &&ViT B-16 + ResNet &99.70& 99.45 &90.97& 99.97
 &96.76 &99.90
&94.42&93.96&94.48&99.01&94.36&98.35&89.09&84.49&\\
 \addlinespace[1mm]
 
  &&N-ResNet + ResNet & 99.74 & 99.54 &87.08&99.97&99.02 &99.91&94.85&94.92&95.64&98.93&96.57&98.53&87.83&80.01&\\
 \addlinespace[1mm]

&&ViT L-32 + ResNet  & 99.69
  & 99.39 &  87.39 & 100  &98.19&99.94&95.22&94.75&95.21&98.70&93.89&97.90&84.02&80.94&\\
 \addlinespace[1mm]

 &&ViT B-16 + ViT L-32 &99.92 & 99.95 &91.57& 99.94 &92.59 &99.88&94.51&92.75&94.69&99.32&89.85&98.02&85.22&84.54&\\
 \addlinespace[1mm]

 &&ViT B-16 + ViT L-32 + N-ResNet& 99.91
 &99.87&94.10& 99.97  &97.47 &99.89 &95.45 &94.90  &96.08  &99.84 &95.58&99.47&90.23&85.61 &\\
 \addlinespace[1mm]

  &&ViT B-16 + ViT L-32 +  EfficientNet & 99.97 
 &99.92&89.60& 99.96   &93.53  &99.90&93.89 &93.35  &94.85  &99.40 &89.21&98.53 &92.30&85.89 &\\
 \addlinespace[1mm]
 
   &&ResNet+ N-ResNet +  EfficientNet & 99.85
 &99.75&87.78& 99.95  &97.91  &99.88&93.89 &94.43  &95.27 &99.36 &94.91&99.12 &92.87&86.57 &\\
 \addlinespace[1mm]
 
  &&ViT B-16 + ViT L-32 +  EfficientNet +ResNet & 99.73
 &99.52&89.74& 99.98 &96.47  &99.91&94.43 &94.32  &95.23 &99.43 &92.42&98.90 &91.98&85.95&\\
 \addlinespace[1mm]

  &&ViT B-16 + ViT L-32 +  EfficientNet + ResNet +N-ResNet  &  99.87&99.78&92.22&99.97&97.71&99.90&94.85&94.99&95.85&
99.65&94.90&99.35&92.69&86.16\\

 \addlinespace[1mm]
 \addlinespace[1mm]
 \addlinespace[1mm]
\specialrule{.07em}{.1em}{.1em}
 \addlinespace[1mm]
  \addlinespace[1mm]
 \multirow{12}{*}{\rotatebox[origin=c]{90}{\makecell{APCER}}}

  &&ViT B-16 + ResNet  & 2.34 & 5.48 &76.41& 0.08 &32.14&0.24&31.75&50.28&35.16&17.84&59.23&29.46&76.84&80.05&\\
  
 \addlinespace[1mm]
  &&N-ResNet + ResNet & 3.03 & 5.14 &68.38& 00.0 &14.89 &00.0&28.73&37.05&28.73&17.53&48.13&26.86&61.84&92.77&\\

 \addlinespace[1mm]
  &&ViT L-32 + ResNet & 5.24 & 11.55 &86.89& 0.16 &8.26 &0.24&33.64&43.85&37.42&23.65&58.50
&35.16&78.44&89.55&\\
 \addlinespace[1mm]
 
  &&ViT B-16 + ViT L-32 & 1.19 & 1.57 &61.75& 0.40 &55.31 &0.73&39.69&60.86&39.50&11.72&73.96&26.76&73.49&74.69&\\
 \addlinespace[1mm]

  &&ViT B-16 + ViT L-32 + N-ResNet& 1.74 & 3.50 &54.62& 0.32 &19.80 &0.65&31.75&46.69&30.62&2.17&51.14&8.92&63.58&74.29&\\
 \addlinespace[1mm]

  &&ViT B-16 + ViT L-32 +  EfficientNet& 0.22 & 0.91 &75.83& 0.24 &52.37 &0.32&38.75&55.95&37.80&9.647&75.93&21.16&70.14&74.02&\\
  
   \addlinespace[1mm]
&&ResNet+ N-ResNet +  EfficientNet& 1.97
 &4.16&77.72& 0.16  &24.87  &0.32&32.51 &39.13  &29.86 &10.06 &49.37&17.21 &61.17& 86.57&\\
 \addlinespace[1mm]

 &&ViT B-16 + ViT L-32 +  EfficientNet +ResNet& 0.09
 &0.41&73.30& 0.16 &29.78  &0.24&34.21 &46.31  &34.59 &7.36 &55.60&16.49 &73.62& 72.28&\\
 \addlinespace[1mm]

 &&ViT B-16 + ViT L-32 +  EfficientNet + ResNet +N-ResNet &1.97&4.30&80.01&0.16&29.21&0.32&33.08&39.5&29.86&3.73&48.44&7.98&58.63&72.28&\\
 \addlinespace[1mm]
 
 \addlinespace[1mm]
\bottomrule
\bottomrule
\end{tabular}%
}

\label{result_2}%
\end{table*}%

\begin{figure}[t] 
\centering
    \includegraphics[width=1\linewidth]{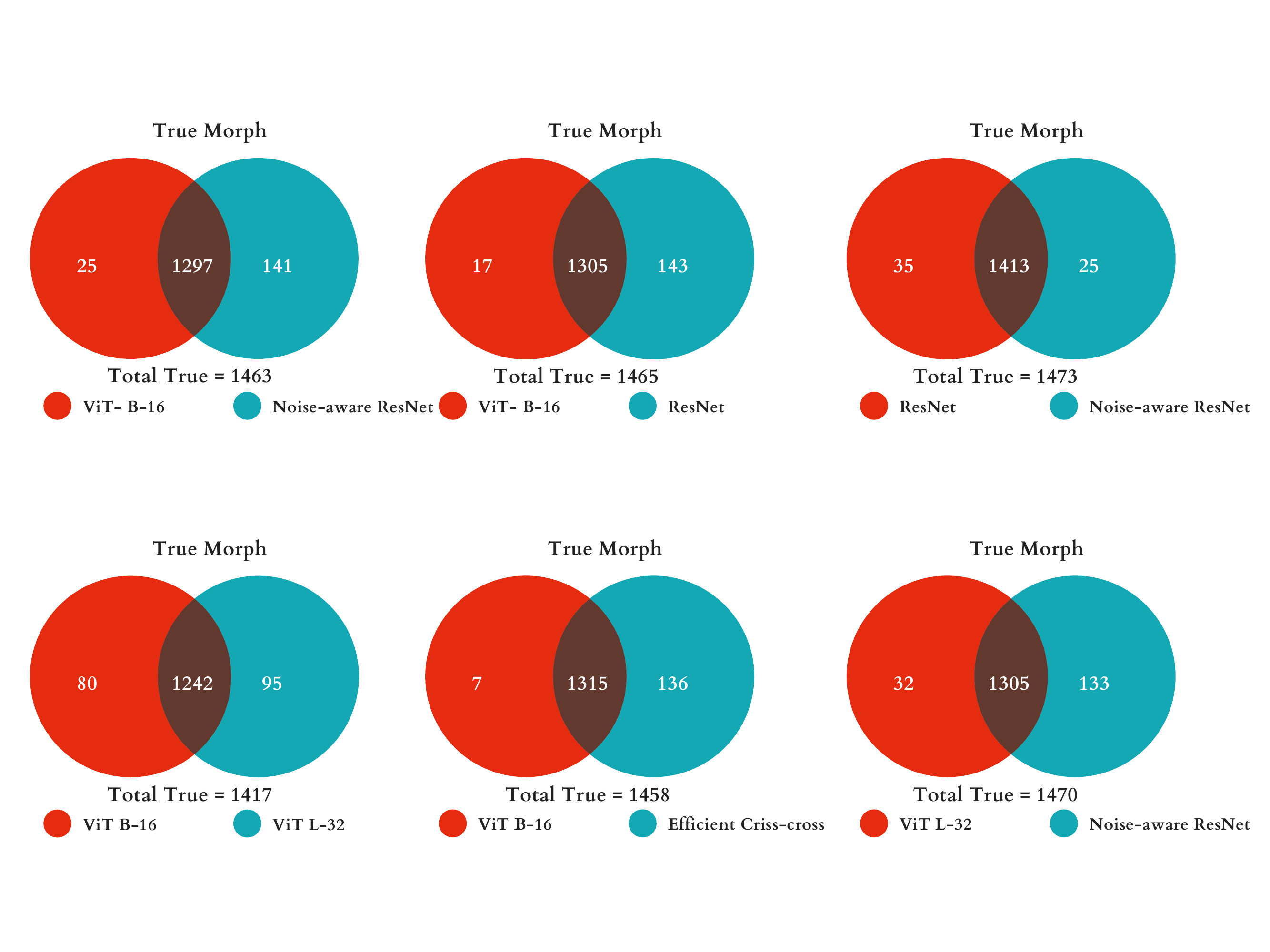}
    \caption{The number of morphs images that are correctly classified as the morph class. }
    \label{venn_fig}
\end{figure}
\paragraph{Single Models:}
To create our ensemble model, several single models based on the CNN and Transformer architectures are trained individually using the source domain. To demonstrate that the error distribution in morph detection is not the same between different models and the proposed ensemble model contributes to the overall performance of morph detection, we apply each single model to the morph images created by the StyleGAN attack \cite{Sarkar2020}. Figure \ref{venn_fig} represents the Venn diagram of morph detection between the ViT B-16, ViT L-32, Inception-ResNet-V1 (abbreviated as ResNet), Noise-aware ResNet, and EffitientNet Criss-cross models. In the EffitientNet Criss-cross model, the Criss-cross attention \cite{huang2019ccnet} is integrated into the EffitientNet architecture \cite{tan2019efficientnet}. The number of true morphs detected by each single model and both of them is provided. The results reveal that the errors are not totally correlated between the single models and they do not perform similarly in different parts of feature space. This demonstrates that the ensemble model, which is composed of different single models, can promote morph detection performance and generalize better to different domains. To select the best models in the ensemble model, we gauge the performance of the single models on 14 different test sets (Table \ref{dataset}). The comparison results in terms of AUC and APCER (@BPCER=0.1) are reported in Table \ref{result_1}. The results show that in some target domains such as \texttt{\#\small{$1$}}, \texttt{\#\small{$2$}}, \texttt{\#\small{$4$}}, and \texttt{\#\small{$6$}}, single models generally perform satisfactorily. However, in most cases such as target datasets \texttt{\#\small{$3$}}, \texttt{\#\small{$5$}}, \texttt{\#\small{$7$}}, \texttt{\#\small{$8$}}, \texttt{\#\small{$9$}}, \texttt{\#\small{$10$}}, \texttt{\#\small{$11$}}, \texttt{\#\small{$12$}}, and \texttt{\#\small{$13$}}, the single models perform differently. For instance, ViT B-16, ResNet, and Noise-aware ResNet models significantly outperform the other single models in target datasets \texttt{\#\small{$3$}}, \texttt{\#\small{$5$}}, \texttt{\#\small{$9$}}, respectively. This analysis indicates that an ensemble model constructed from these single models could potentially gain noticeable improvements in comparison with the single models, mainly {for} the challenging target domains.

\begin{table*}[t]
\center
\caption{ Morph detection results for the selected ensemble model (ViT B-16 + ViT L-32 + N-ResNet) for different fusion strategies. The evaluations are in terms of APCER (@BPCER=1\%) and AUC metrics. }
\resizebox{.99\textwidth}{!}{%
\begin{tabular}{ccccccccccccccccccc}
\toprule
\toprule

   && Methods  & {\rotatebox[origin=c]{60}{\makecell{AMSL} }}& 
   {\rotatebox[origin=c]{60}{\makecell{FRLL \\AMSL} }}& 
   {\rotatebox[origin=c]{60}{\makecell{FRLL\\ Webmorpher  }}}&
   {\rotatebox[origin=c]{60}{\makecell{FRLL \\OpenCV }}}&
   {\rotatebox[origin=c]{60}{\makecell{FRLL \\StyleGAN}}}&
   {\rotatebox[origin=c]{60}{\makecell{FRLL\\ Facemorpher }}}&
   {\rotatebox[origin=c]{60}{\makecell{FERET\\ OpenCV}}}& 
   {\rotatebox[origin=c]{60}{\makecell{FERET\\ StyleGAN}}}&
   {\rotatebox[origin=c]{60}{\makecell{FERET\\ Facemorpher}}}&
   {\rotatebox[origin=c]{60}{\makecell{FRGC \\OpenCV}}}& 
   {\rotatebox[origin=c]{60}{\makecell{FRGC \\StyleGAN}}}&
   {\rotatebox[origin=c]{60}{\makecell{FRGC \\Facemorpher}}}&
   {\rotatebox[origin=c]{60}{\makecell{FRGC \\MIPGAN}}}&
   {\rotatebox[origin=c]{60}{\makecell{FRGC \\MIPGAN\\ (PRINT)}}}\\ \midrule
 \addlinespace[1mm]
 \addlinespace[1mm]

  \multirow{6}{*}{\rotatebox[origin=l]{90}{\makecell{AUC}}}

 && Soft Voting& 99.91
 &99.87&94.10& 99.97  &97.47 &99.89 &95.45 &94.90  &96.08  &99.84 &95.58&99.47&90.23&85.61 &\\
 \addlinespace[1mm]

 &&Max Voting 
 &99.74&99.63& 88.92  &99.96 &97.77 &99.86 &95.71  &95.32  &96.29 &99.83&95.08&99.37&88.30&85.11&\\
 
 \addlinespace[1mm]

 && Score-based Super Learner&   99.81 &99.73&93.22&99.96&97.82&99.87&95.36&94.99&96.40&99.86&96.73&99.63&91.29&86.75 &\\
 \addlinespace[1mm]

 && Feature-based Super Learner& 99.91&99.83&92.08&99.98
 &98.08
 &99.89
 &95.83
 &95.78
 &96.70
 &99.78
 &96.48
 &99.69
 &91.86
 &85.81 &\\

 \addlinespace[1mm]
\specialrule{.07em}{.1em}{.1em}
 \addlinespace[1mm]

 \multirow{5}{*}{\rotatebox[origin=c]{90}{\makecell{APCER}}}

 &&Soft Voting & 1.74 & 3.50 &54.62& 0.32 &19.80 &0.65&31.75&46.69&30.62&2.17&51.14&8.92&63.58&74.29&\\
 \addlinespace[1mm]


 &&Max Voting&1.93
 &3.69&53.89& 0.32  &20.45 &0.65 &31.37 &46.69 &30.62  &2.17 &51.65&9.85&63.58&74.29 &\\
 
 \addlinespace[1mm]

 &&Score-based Super Learner&3.63&5.08&62.98& 0.32 &29.78 &1.22 &31.19&46.88 &30.43   &1.45&44.19&5.39&56.35&69.61 &\\
 \addlinespace[1mm]

 &&Feature-based Super Learner&1.70&3.21&60.19& 0.16
 &26.43
 &0.40 
 &28.16
 &40.26
 &24.76
 &1.97
 &47.09
 &3.42
 &55.55
 &72.04 &\\
 \addlinespace[1mm]

 \addlinespace[1mm]
\bottomrule
\bottomrule
\end{tabular}%
}

\label{result_3}%
\end{table*}%

\subsection{Ensemble Model}
To find the best configuration of the ensemble model, we conduct two ablation studies as in Table \ref{result_2}. In the first study, the number of single models, as well as their combinations, are ablated in the ensemble model. The soft voting strategy is adopted to fuse different single models. More specifically, the probability scores of the single models are averaged in the ensemble model. For the sake of brevity, only the best combinations are reported in Table \ref{result_2}. It is observed that naively averaging the probability scores of the single models could generally boost the generalization of the ensemble model to different target domains. This is largely ascribed to the lower variance in the ensemble models compared to the single models. In addition, by drawing a comparison between different configurations of the ensemble model, we {observe} that the ViT B-16 + ViT L-32 + N-ResNet \cite{dosovitskiy2020image,venkatesh2020detecting} combination generally yields the best performance among the ensemble models with two and three components. This evaluation also point{s} out that the larger number of single models would not necessarily boost the generalization of the final ensemble model, and {may} only incurs more computational costs.

\begin{table}

\center
\caption{Comparison of the selected ensemble model and the robust ensemble model with adversarial training. The robust accuracy against different adversarial attacks is in terms of AUC/APCER@BPCER=1\%  metrics on a subset of FRGC MIPGAN. In black-box attacks, stronger perturbation levels are utilized to generate adversarial perturbations in comparison to white-box attacks.}

\begin{adjustbox}{max width=0.48\textwidth}
\begin{tabular}{ccccccccc}
\toprule
\toprule

 \addlinespace[1mm]
 && Target &  DIFGSM \cite{xie2019improving}& MIFGSM \cite{dong2018boosting}& TIFGSM \cite{dong2019evading} &TPGD \cite{zhang2019theoretically} &  Square \cite{andriushchenko2020square} &  C\&W \cite{7958570}\\
  \addlinespace[1mm]
 \midrule
 \addlinespace[4mm]

  \multirow{2}{*}{\rotatebox[origin=l]{90}{\makecell{White-Box}}}

 &&Ensemble Model
 &84.60 / 93.75 &80.67 / 96.875 & 83.26 / 95.08 &71.39 / 97.32 & 74.17 / 100 & 74.80 / 100\\
 \addlinespace[3mm]

 && Robust Ensemble Model
 &88.87 / 87.05 &91.82 / 59.37 & 89.43 / 84.87  &96.22 / 30.35 & 94.04 / 59.37 & 91.82 / 59.37\\

 \addlinespace[6mm]
\specialrule{.07em}{.1em}{.1em}
 \addlinespace[4mm]
  
  \multirow{3}{*}{\rotatebox[origin=l]{90}{\makecell{Black-Box}}}

 &&Ensemble Model
 &32.0 / 100 &49.9 / 99.5& 15.2 / 100 &80.4 / 44.6 & 91.8 / 78.1& 86.8 / 93.7\\
 \addlinespace[3mm]

 && Robust Ensemble Model
 &98.0 / 28.1 & 97.6 / 30.8 & 97.4 / 47.3  & 98.6 / 19.2 & 92.9 / 54.9& 91.5 / 59.8\\

 \addlinespace[5mm]
\bottomrule
\bottomrule
\end{tabular}%

   \end{adjustbox}
\label{result_4}%
\end{table}%

\begin{table}

\center
\caption{Comparison of the selected robust ensemble model with state-of-the-art morph detection models in the FRLL and the LMA-DRD \cite{damer2021pw} datasets. The results are in terms of EER, and BPCER (@APCER=1\% and 10\%) metrics.}

\resizebox{.46\textwidth}{!}{%
\begin{tabular}{ccccc}
\toprule
\toprule
\addlinespace[1mm]

 &Target &  D-EER & BPCER (1\%)  & BPCER (10\%)    \\ 
 \midrule
  \multirow{7}{*}{\rotatebox[origin=l]{90}{\makecell{FRLL}}}
 &MixFacenet - SMDD \cite{damer2022privacy}
 &3.87 &23.53  & 0.49 \\
 \addlinespace[1mm]

 &PW-MAD - SMDD \cite{damer2022privacy}
 &2.20  &26.47  & 0.49 \\
 \addlinespace[1mm]

 &Inception - SMDD \cite{damer2022privacy}
 &3.17  &30.39  & 0.49\\
  
  \addlinespace[1mm]
  &Denoising based method  \cite{venkatesh2020detecting}
 &1.96 &5.39& 00.0   \\
 
  \addlinespace[1mm]
 
  & Ensemble Model
 &0.98 &0.98& 00.0   \\
 
 \addlinespace[1mm]
 
   &Robust Ensemble Model
 &0.98 &0.98& 00.0   \\
 \midrule
   \multirow{5}{*}{\rotatebox[origin=l]{90}{\makecell{LMA-DRD}}}

 &MixFacenet - SMDD \cite{damer2022privacy}
 &19.42 &79.67  & 31.71 \\
 \addlinespace[1mm]

 &PW-MAD - SMDD \cite{damer2022privacy}
 &17.06  &80.49  & 25.20 \\
 \addlinespace[1mm]

 &Inception - SMDD \cite{damer2022privacy}
 &15.11 &69.92  & 30.89\\

  \addlinespace[1mm]
 
  & Ensemble Model
 &13.64 &72.35&17.07  \\

\bottomrule
\end{tabular}%
}

\label{result_5}%
\end{table}%

In the second study, different fusion strategies are ablated in Table \ref{result_3}. The fusion strategies include soft voting, feature-based super learner, and score-based super learner strategies. In our score-based super learner, we freeze the single models and train another learner, which is a single-layer FFN, to weigh the output scores of the single models. Also, in our feature-based super learner, the output features from the last FC layers of different single models are concatenated and fed to a two-layer FFN for final morph classification (Figure \ref{fig2}]). Since the ViT B-16 + ViT L-32 + N-ResNet model relatively outperforms the other configurations in Table \ref{result_3}, we set them as the components of the final selected ensemble model. The comparisons indicate that the feature-based super learner outperforms other fusion strategies to some extent. {From this experiment}, we can deduce that the ensemble model with ViT B-16, ViT L-32, N-ResNet, and the feature-based super learner components outperforms its competitor ensemble models. 

\subsection{Robustness}

In this section, the robust accuracy of the selected ensemble model is targeted. For this objective, multi-perturbation adversarial training is employed. To benchmark the effectiveness of the model-based ensembling attack \cite{liu2017delving} in the adversarial training, we gauge the robustness of the ensemble model that has been adversarially trained. In the robustness evaluations, we utilize new adversarial attacks of which the ensemble model is not aware. Based on Table \ref{result_4}, the performance of the ensemble model against unseen adversarial attacks in the white-box and black-box settings drops significantly. It is observed that the robust ensemble model gains substantial improvements over the baseline ensemble model against DIFGSM \cite{xie2019improving}, MIFGSM \cite{dong2018boosting}, TIFGSM \cite{dong2019evading}, TPGD \cite{zhang2019theoretically}, Square \cite{andriushchenko2020square}, and  C\&W \cite{7958570} attacks in both white-box and black-box settings. In short, taking these results into account, we can substantiate that the multi-perturbation adversarial training with the model-based ensembling optimization \cite{liu2017delving} improves the robust accuracy of our final model against several adversarial attacks in white-box and black-box settings. In the last experiment, we make a comparison between the proposed robust ensemble model and various state-of-the-art studies. This experiment is also carried out to find out whether the multi-perturbation adversarial training detrimentally hurts the clean accuracy of the ensemble model or not. The results in Table \ref{result_5} demonstrate that the proposed robust ensemble model maintains its superior performance on clean accuracy and also significantly surpasses the state-of-the-art studies.

\section{Conclusions}

In this paper, we present a morph attack detection with strong generalization ability to different morph attacks and high robustness against adversarial attack. By combining CNN and Transformer models in our ensemble model, we capture both long-term and short-term relationships in morph images to improve domain generalization. To strengthen the robustness of our model against adversarial attacks, we employ multi-perturbation adversarial training with highly transferable adversarial examples. Experimental results on several datasets demonstrate the generalization ability and adversarial robustness of our proposed model. 

\paragraph{Acknowledgements.}This material is based upon a work supported by the Center for Identification Technology Research and the National Science Foundation under Grant \#1650474.

\newpage

{\small
\bibliographystyle{ieee}
\bibliography{paperArxiv}
}

\end{document}